\documentclass{article}

\usepackage{microtype}
\usepackage{graphicx}
\usepackage{subfigure}
\usepackage{booktabs} 

\usepackage{hyperref}

\usepackage[accepted]{icml2024}

\usepackage{amsmath}
\usepackage{amssymb}
\usepackage{mathtools}
\usepackage{amsthm}

\usepackage[capitalize,noabbrev]{cleveref}

\theoremstyle{plain}

\theoremstyle{definition}

\theoremstyle{remark}

\usepackage[textsize=tiny]{todonotes}

\begin{document}

\twocolumn[
\icmltitle{Tackling Polysemanticity with Neuron Embeddings}



\icmlsetsymbol{equal}{*}

\begin{icmlauthorlist}
\icmlauthor{Alex Foote}{ripjar,apart}
\end{icmlauthorlist}

\icmlaffiliation{ripjar}{Ripjar}
\icmlaffiliation{apart}{Apart Research}

\icmlcorrespondingauthor{Alex Foote}{alexjfoote@icloud.com}

\icmlkeywords{Machine Learning, ICML, Mechanistic Interpretability}

\vskip 0.3in
]



\printAffiliationsAndNotice{}  

\begin{abstract}
We present neuron embeddings, a representation that can be used to tackle polysemanticity by identifying the distinct semantic behaviours in a neuron's characteristic dataset examples, making downstream manual or automatic interpretation much easier. We apply our method to GPT2-small, and provide a UI for exploring the results. Neuron embeddings are computed using a model's internal representations and weights, making them domain and architecture agnostic and removing the risk of introducing external structure which may not reflect a model's actual computation. We describe how neuron embeddings can be used to measure neuron polysemanticity, which could be applied to better evaluate the efficacy of Sparse Auto-Encoders (SAEs).
\end{abstract}

\section{Introduction}
\label{intro}

Mechanistic Interpretability (MI) aims to decompose neural networks into their constituent parts and understand how these parts interact to create the behaviour of the network, with the ultimate goal of understanding models in enough detail to determine whether they're safe to deploy. One of the key suppositions of MI is that it's possible to break models apart into meaningful units, often called features \cite{circuits}. One natural basis for these units is the neuron, and visualisation techniques developed for vision models had significant success in understanding the function of many neurons, as well as how they compose to implement increasingly complex behaviours \cite{curvedetectors}. 

However, a major obstacle to this approach is the fact that neurons often respond to several completely distinct concepts, a phenomenon called polysemanticity. This makes it much harder to find a clean and simple explanation for a neuron's behaviour, and undermines the idea that neurons are the natural basis for decomposing a model. Polysemanticity is particularly prevalent in language models, and has made interpreting their MLP layers a significant challenge \cite{solu}.

One common method for interpreting the behaviour of a neuron in a language model is to collect and study the dataset examples which cause the highest neuron activation. Patterns in a neuron's dataset examples provide an indication of what the neuron responds to. However, polysemanticity makes these dataset examples much harder to interpret, as there are often many separate behaviours to understand, some of which may be related and others entirely distinct. This becomes increasingly challenging as you collect examples further down the activation spectrum, which is important for gaining a complete understanding of a neuron, but often reveals a wider range of behaviours \cite{illusion}.

To tackle the problem of polysemanticity, we introduce \textbf{neuron embeddings}, which capture the information that a given neuron is responding to in a given input. Given a neuron which we're trying to understand and an input which causes that neuron to activate, we define the neuron embedding of the input as the element-wise product of the vector representation that the neuron receives and the neuron's input weights. We show that this representation can be used to cluster a neuron's dataset examples, making it possible to disentangle the neuron's behaviour into it's constituent parts. Dataset examples are used for both manual and automated interpretability \cite{auto_interp, n2g}, so making them easier to interpret has significant potential benefit for a variety of downstream applications. We apply this method to GPT2-small \cite{GPT2} and provide case studies on individual neurons, as well a website for exploring the results for the full model \footnote{\url{https://feature-clusters.streamlit.app}}.

Crucially, neuron embeddings also allow us to measure a proxy for a neuron's degree of polysemanticity by computing simple metrics on the geometry of the points and the clusters that form. Sparse Auto-Encoders (SAEs) \cite{saes} are a promising technique for dealing with polysemanticity, which learn to disentangle a layer of neurons into a wider, sparse MLP layer with monosemantic neurons. However, we lack effective metrics for evaluating the quality of SAEs, instead relying on simple heuristics like reconstruction error and activation sparsity, as well as time consuming manual analysis. Neuron embeddings may be able to bridge this gap by providing better metrics for measuring polysemanticity of the SAE neurons, which typically corresponds well with neuron interpretability.

Finally, we also provide a proof-of-concept demonstrating how we can incorporate neuron embeddings into the training of SAEs, by computing a measure of neuron monosemanticy and including it in the SAE loss. We show how this affects SAEs for a toy MLP trained on MNIST \cite{mnist}, decreasing the reconstruction accuracy and activation sparsity but increasing monosemanticity and significantly decreasing the prevalence of dead neurons which never activate.

\section{Related Work}

The favoured explanation for the cause of polysemanticity is superposition, which supposes that models learn to encode features across multiple neurons, allowing them to represent many more features than they have neurons. Superposition has been compellingly demonstrated in toy models \cite{toysuperposition} and large language models \cite{gurnee2023finding}.

One thread of research aimed to tackle superposition and polysemanticity by introducing a new activation function, called a Softmax Linear Unit (SoLU) \cite{solu}, which encourages neurons to activate sparsely. However, they found that this resulted in polysemanticity effectively being pushed into other neurons and the Layer Norm component, rather than truly eliminated.

Dictionary learning is an alternative and very promising method for interpreting language models in spite of superposition and polysemanticity. It aims to decompose a model's internal embeddings or MLP layers into features, by training a Sparse Auto-Encoder (SAE) layer to reconstruct their activations. SAEs have been explored by a number of groups \cite{saes, cunningham2023sparse, yun2023transformer}, most notably in a recent paper that applied them to the residual stream of a very large production language model and found a rich array of interpretable features \cite{sonnet}.

\section{Method}
\label{method}

\begin{figure*}
\centering
\includegraphics[width=\textwidth]{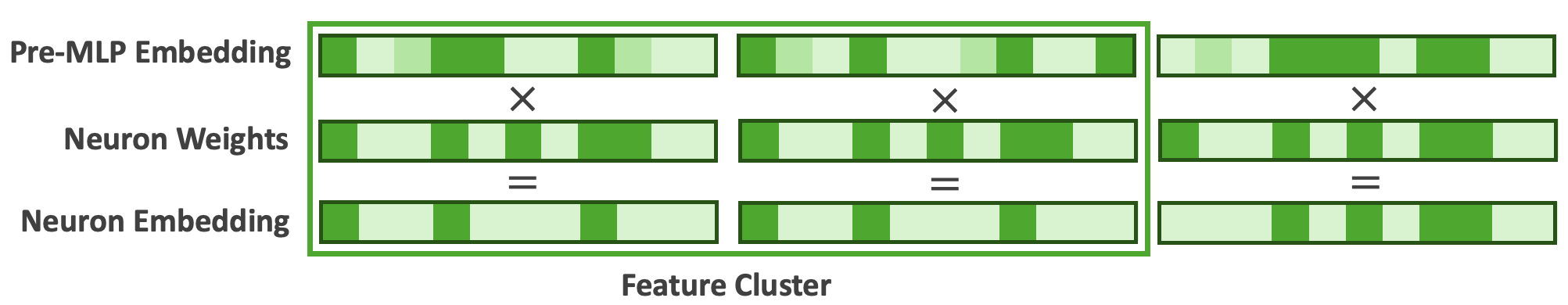}
\vskip -0.1in
\caption{An illustration of the neuron embedding process. We compute the element-wise products of the pre-MLP embedding of the inputs and the neuron input weights to produce the neuron embeddings. These are then clustered based on similarity. The neuron weights select the relevant information from the embedding, such that the neuron embeddings of two different inputs can be brought together or pushed apart.}
\label{figure:diagram}
\end{figure*}

\subsection{Neuron Embeddings}

Given a model containing $l$ (potentially non-contiguous) MLP layers, we denote the $i$th layer as $L_i$, and the $j$th neuron in $L_i$ as $N_{i,j}$, which has input weights $\mathbf{w}_{i,j}$. For a given model input $\mathbf{x}$, we denote the internal vector representation of $\mathbf{x}$ immediately before layer $L_i$ as $\mathbf{h}_{i-1}$, which we refer to as the pre-MLP embedding.

We then define the neuron embedding of $\mathbf{x}$ for the neuron $N_{i,j}$ as $\mathbf{e}_{i,j}$, where $\mathbf{e}_{i,j}$ is the Hadamard product of the pre-MLP embedding $\mathbf{h}_{i-1}$ and the input weights of the neuron $\mathbf{w}_{i,j}$:
\begin{equation}
    \mathbf{e}_{i,j} = \mathbf{h}_{i-1} \odot \mathbf{w}_{i,j}
\end{equation}
This is exactly the first stage of computing the neuron's activation - summing $\mathbf{e}_{i,j}$ and applying the activation function would complete the process. Constraining ourselves to use a representation that is computed internally to the model reduces the risk of introducing additional structure which may not be true to what the model is actually doing. It also means neuron embeddings can be computed for any MLP neurons, regardless of the rest of the model architecture (CNNs, Transformers, etc.) or the domain (vision, language, etc.).

Intuitively, the input weights of a neuron represent what the neuron is ``looking for" in an input - with different parts of the weights potentially looking for different features. The neuron embedding then, in some sense, represents the feature that the neuron found in the input that caused it to activate. If this does indeed produce a good representation of the feature which caused activation, we should then be able to use it to separate the mixture of dataset examples into their distinct semantic behaviours. Figure \ref{figure:diagram} illustrates this process.

\subsection{Feature Clusters}

We now apply neuron embeddings to tackle the problem of polysemantic neurons by creating \textbf{feature clusters}. A feature cluster is a set of highly activating dataset examples that capture a single semantic behaviour of a neuron, created by clustering the neuron embeddings of a neuron's dataset examples. We demonstrate how feature clusters can be computed for autoregressive GPT-style language models, but note that this technique can be applied to any model containing MLP layers. 

Extending the above notation, we define the $k$th dataset example for a neuron $N_{i,j}$ as $\mathbf{x}_{i,j,k}$, which consists of a sequence of tokens $t = t_1 \cdots t_n$ where $N_{i,j}$ strongly activates on the last token $t_n$. The neuron embedding $\mathbf{e}_{i,j,k}$ of the dataset example $\mathbf{x}_{i,j,k}$ for neuron $N_{i,j}$ is therefore:
\begin{equation}
    \mathbf{e}_{i,j,k} = \mathbf{h}_{i-1,k,n} \odot \mathbf{w}_{i,j},
\end{equation}
where $\mathbf{h}_{i-1,k,n}$ is the pre-MLP embedding of the activating token $t_n$.

Given a set of highly activating dataset examples for a target neuron, we compute the neuron embeddings for each example. We then compute the pairwise distance matrix between all pairs of embeddings, where the distance $d$ between dataset examples $k_a$ and $k_b$ is:

\begin{equation}
    d = 1 - S_C(\mathbf{e}_{i,j,k_a}, \mathbf{e}_{i,j,k_b}),
\end{equation}

where $S_C(\cdot)$ is the cosine similarity. We use Hierarchical Agglomerative Clustering (HAC) on the distance matrix with a distance threshold that controls when clusters merge, which we set to $0.5$. HAC also provides a full cluster hierarchy, which can give additional insight into a neuron's behaviour. In particular, neurons may have some semantically related but distinct features, as well as completely unrelated features. The hierarchy clearly captures these relationships between clusters, as well as between sub-clusters.

\subsection{Monosemantic Training}

\subsubsection{Measuring Polysemanticity}

By computing the neuron embeddings for a neuron's dataset examples, we can naturally compute metrics on the embedded points, allowing us to measure proxies for monosemanticity. We compute simple metrics like the maximum distance between any pair of points, the mean distance between points, and metrics on the clustering such as mean inter- and intra-cluster distance, and the number of clusters.

We can also apply this method to the neurons in Sparse Auto-Encoders (SAEs), which are trained to disentangle the features represented in MLP layers. Evaluating SAEs currently relies on a few heuristics that tend to correlate with interpretable, monosemantic neurons, such as the average number of SAE neurons activating per example, in combination with manual interpretation of neurons \cite{saes}. These neuron embeddings metrics more directly measure neuron monosemanticity, and so would be a natural addition to SAE evaluation to bridge the gap between fast but loosely-correlated heuristics, and slow manual interpretation.

\subsubsection{Sparse Auto-Encoder Loss}

When training SAEs, we want each SAE neuron to represent a single, interpretable feature. This is achieved by pushing the neurons to activate sparsely by including the $L1$ norm of the neuron activations in the loss. However, this does not directly penalise neurons for responding to multiple features. We show that we can utilise the ability of neuron embeddings to capture the similarity between features to directly penalise polysemantic neurons via a term in the SAE loss.

Intuitively, we want to push neurons to respond to a single feature, which corresponds to a neuron having a single dense cluster in it's neuron embedding space. To do this efficiently during training, for each SAE neuron we maintain a neuron embedding of the inputs which activated the neuron, and then convert this to a neuron embedding on the fly to measure the similarity of the combined neuron embedding with the neuron embedding of the current input. We then take the sum of these similarities for all neurons over a batch of inputs and include it in the SAE loss.

Assume we're training an SAE layer with dimensionality $d$, inserted after layer $L_i$ of the original model. Concretely, at a given training step, for each input $\mathbf{x}_b$ in a batch of size $n$, $\{ \mathbf{x}_0 \cdots \mathbf{x}_{n-1} \}$, we collect all the SAE neurons with a non-zero activation. For each neuron, we check if it's activated before. If it hasn't, we store the pre-SAE embedding $h_i$ in a lookup mapping the neuron to an embedding. If it has, we retrieve the averaged embedding $h_{i,j,avg}$ stored in the lookup and convert it to a neuron embedding $\mathbf{e}_{i,j,avg}$ by taking it's Hadamard product with $\mathbf{w}^{SAE}_{i,j}$, the weights of SAE neuron $N_{i,j}^{SAE}$. We similarly compute the neuron embedding $\mathbf{e}_{i,j,b}$ of the current input $\mathbf{x}_b$ which has caused the neuron to activate, using it's pre-SAE embedding $h_{i,b}$. 

We then compute the distance between the two neuron embeddings, $1 - S_C(\mathbf{e}_{i,j,avg}, \mathbf{e}_{i,j,b})$, which measures whether the feature the neuron is responding to in the current input is similar to the average feature the neuron has responded to throughout training. We take the sum of the distances across all neurons and all inputs in the batch to compute the neuron embedding loss $\mathcal{L}_{N}$. In a single equation:

\begin{equation}
    \mathcal{L}_{N} = \sum_{b=0}^n \sum_{j=0}^d 1 - S_C(h_{i,j,avg} \odot \mathbf{w}^{SAE}_{i,j}, h_{i,b} \odot \mathbf{w}^{SAE}_{i,j})
\end{equation}

We then incorporate this into the standard SAE loss $\mathcal{L}_{SAE}$, weighted by $\lambda_2$:

\begin{equation}
    \mathcal{L}_{SAE} = MSE(a_i, a^{SAE}_i) + \lambda_1 \cdot \| a^{SAE}_i \|_1 + \lambda_2 \cdot \mathcal{L}_{N},
\end{equation}

where $MSE$ is the mean-squared error, $a_i$ denotes the MLP activations of $L_i$ and $a^{SAE}_i$ denotes the reconstructed activations from the SAE.

To update the average embedding $h_{i,j,avg}$ for a neuron $N^{SAE}_{i,j}$ across training, we use a momentum-based update:

\begin{equation}
    h_{i,j,avg} = m \cdot h_{i,j,avg} + (1 - m) \cdot h_{i,b},
\end{equation}

where $m$ controls the balance between the existing embedding and the new embedding. This allows information from earlier embeddings to decay, putting more weight on the more recent embeddings which should better reflect what the neuron is learning to respond to. We set $m$ to $0.9$ in our experiments.

Intuitively, this new loss term pushes each neuron to respond to a single feature, as the distance between the combined neuron embedding and the current neuron embedding will increase if the combined neuron embedding contains multiple features.

\section{Results}

\subsection{Feature Clusters}

\begin{figure*}
\centering
\includegraphics[width=\textwidth]{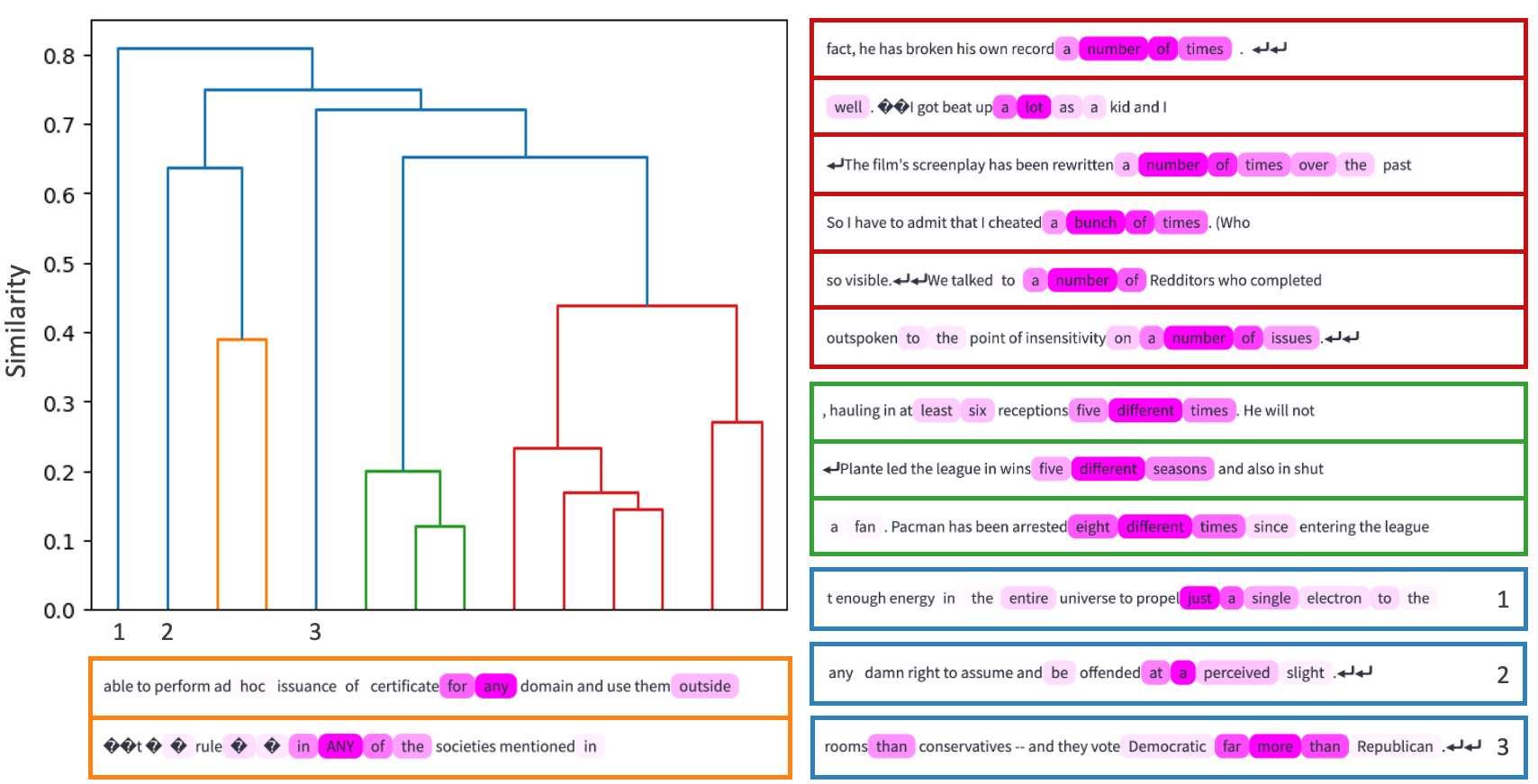}
\vskip -0.1in
\caption{An example of feature clustering applied to a neuron in layer 7 of GPT2-small. The clusters (colour and numerically coded) each show a distinct semantic behaviour, and the dendrogram shows how the cluster hierarchy formed. The highlighting corresponds to neuron activation on each token, with the neuron embedding derived from the maximally activating token.}
\vskip -0.1in
\label{figure:clusters}
\end{figure*}

\subsubsection{Experimental Setup}

We show that neuron embeddings can effectively capture semantic similarity between a neuron's dataset examples by applying feature clustering to GPT2-small. We processed 10,000 input examples from OpenWebText \cite{openwebtext} with the model and retrieved the model's internal activations using the TransformerLens library \cite{transformerlens}. For each neuron, we collected any example that induced at least 75\% of the maximum observed activation of the neuron \footnote{Maximum activation was obtained from Neuroscope \cite{neuroscope}, which measured neuron activation on a much larger dataset}, up to a maximum of 100 examples per neuron. We chose to keep the first 100 examples over this activation threshold, rather than the top 100 most activating examples, to better capture the diversity of behaviours that occur below maximal activation. A basic implementation of this process took a few hours to run on a single GPU, but could likely be made dramatically more efficient with more engineering effort, to enable it to scale to larger models. 

We then applied feature clustering to each neuron to group the dataset examples into their distinct features. We analyse some examples of neurons to demonstrate the efficacy of the technique, and also provide a UI for navigating the full model \footnote{\url{https://feature-clusters.streamlit.app/}}. The UI additionally provides insight into which tokens were important for neuron activation using the method from \cite{n2g}, as well as links to similar features from other neurons identified via Nearest Neighbour search on the embeddings of the central example in each feature cluster.

\begin{figure*}[t]
\centering
\includegraphics[width=\textwidth]{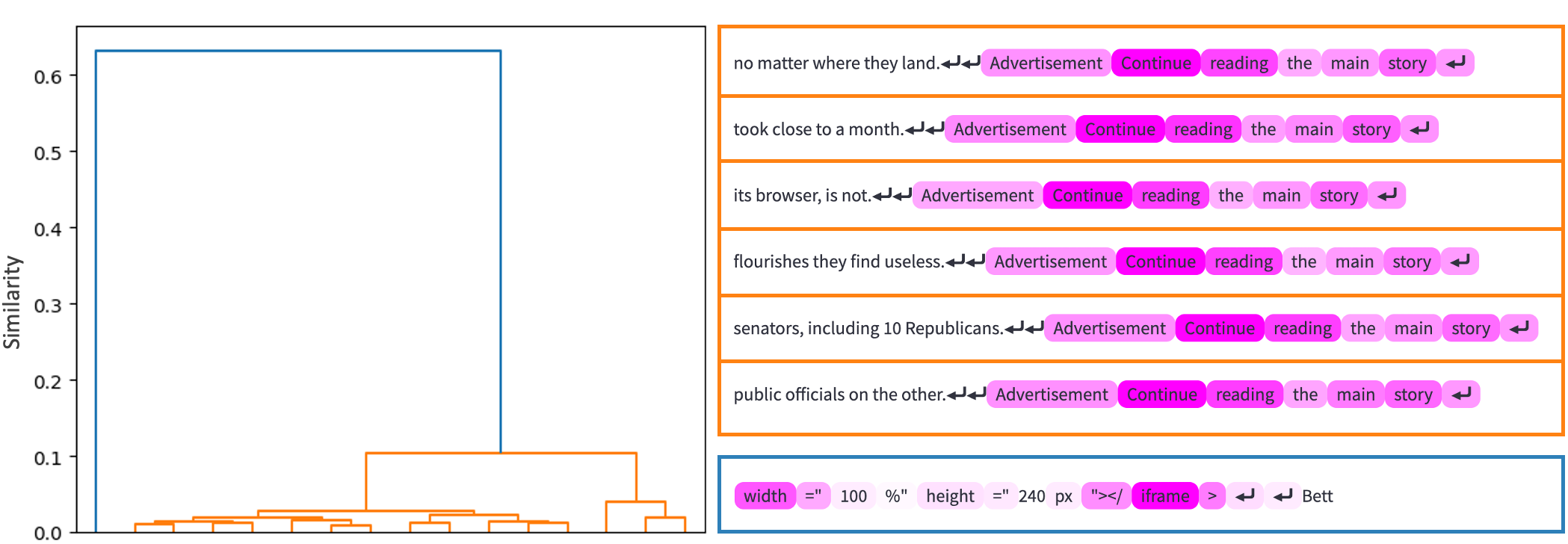}
\vskip -0.1in
\caption{An example of a neuron with a common primary behaviour (orange) and a rare secondary behaviour (blue).}
\vskip -0.1in
\label{figure:split_clusters}
\end{figure*}

\subsubsection{Neuron Examples}


Figure \ref{figure:clusters} shows the results of applying clustering to a neuron in the $7$th layer (of $12$) of the model. The dendrogram shows the results of the hierarchical clustering of the neuron embeddings, where each branch point shows the average distance between the embeddings in the two clusters (where a cluster contains one or more points). The examples separate into six clusters (colour and numerically coded), with snippets of the dataset examples in each cluster shown. The highlighting indicates the strength of neuron activation on each token, and the brightest token is the key token which we use to produce the neuron embedding. Note we also show the following five tokens after the key token for context, but these do not influence the neuron embedding as the model is auto-regressive.

The clusters each contain a distinct behaviour, and successfully group examples that represent the same concept with different wording. The hierarchy captures similarity at multiple levels, with highly similar examples with the same key token clustering first, with these sub-clusters then merging into the full clusters. Additionally, similar but distinct clusters, such as the red and green cluster which both relate to numbers of events, also appear closer together in the hierarchy. 

This phenomenon, where a neuron has multiple related but distinct behaviours, is quite common in GPT2-small, and may be related to feature splitting \cite{saes}, where a neuron in a Sparse Auto-Encoder (SAE) represents multiple closely related features, which then split into increasingly fine-grained features as the size of the Auto-Encoder is scaled up. Applying feature clustering to neurons in SAEs may be able to identify which neurons contain multiple sub-features which might split after scaling up, which could be useful for measuring the prevalence of these neurons and better understanding feature splitting more generally.

Figure \ref{figure:split_clusters} shows an example of a different class of neuron, where there is a common primary behaviour and a rarer secondary behaviour. In this case, dataset examples for the primary behaviour (orange) outnumber those for the rare behaviour (blue) almost 50:1 \footnote{The dendrogram doesn't show the full hierarchy for simplicity - some of the orange leaves are actually clusters with multiple elements}. Without feature clustering, it's very easy to miss these rarer behaviours during interpretation - unless you review a large number of examples even after there appears to be a clear hypothesis for the behaviour, you would naturally conclude that this is a mono-semantic neuron. Feature clustering allows us to collect a much larger number of examples and automatically condense them, making it much easier to quickly identify all the relevant behaviours of an neuron. This could help to address the illusion of interpretability \cite{illusion}, where examining the top examples for a neuron suggests a simple explanation of the behaviour, but expanding to lower activating examples reveals an array of hidden behaviours.

\subsubsection{Comparison to Embeddings}

\begin{figure*}[t]
\centering
\includegraphics[width=\textwidth]{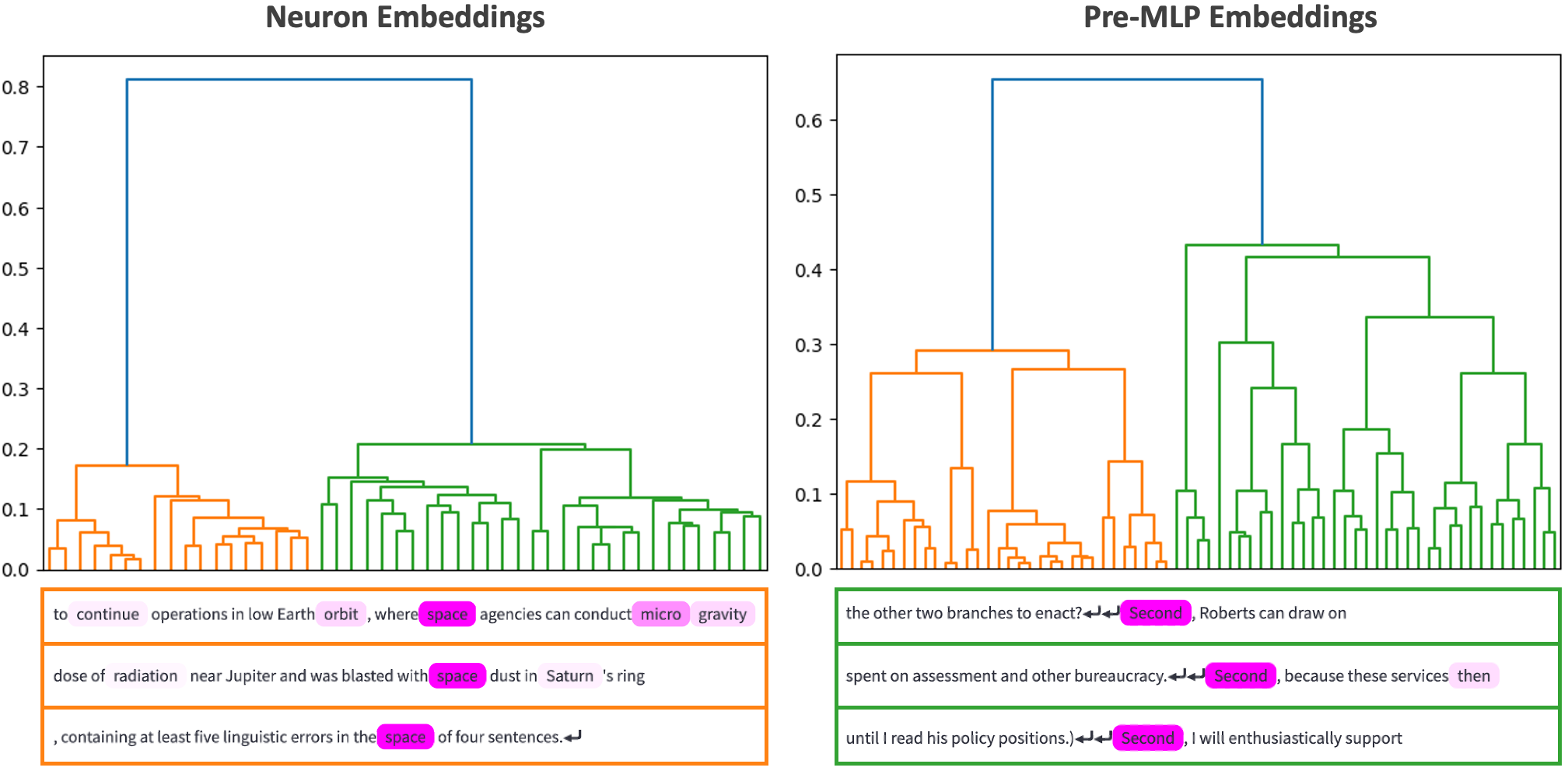}
\vskip -0.1in
\caption{A comparison between feature clusters derived from neuron embeddings vs pre-MLP embeddings. The neuron embeddings clearly result in denser clusters with better separation between the clusters. Examples from the two clusters are shown in their corresponding colours.}
\label{figure:comparison}
\end{figure*}

Whilst we choose to use neuron embeddings to cluster a neuron's dataset examples, we could instead just use the pre-MLP embedding of the key token, without then multiplying it by the neuron's weights. Table \ref{table:distances} compares the median intra- and inter-cluster distances of the dataset example clusterings for all neurons in GPT2-small when using the pre-MLP embeddings or neuron embeddings. It shows that neuron embeddings lead to denser clusters with reduced intra-cluster distance, with better seperation between these clusters from the increased inter-cluster distance. Figure \ref{figure:comparison} clearly illustrates this, showing a pair of clusters with significantly higher density and separation using neuron embeddings compared to pre-MLP embeddings.

The improved separation between clusters implies that feature clusters derived from neuron embeddings will have fewer errors than those derived from pre-MLP embeddings. Anecdotally, we found this to be the case, particularly when using simpler but faster clustering algorithms such as the Sub-Cluster Component algorithm \cite{scc}.

The better performance of neuron embeddings also indicate that they better capture the similarity between a neuron's dataset examples. Intuitively, this is because a neuron may not respond to all information in the pre-MLP embedding, so by incorporating the neuron weights, which represent what the neuron is ``looking for" in an input, we select out the relevant information to the neuron, providing a better representation of what caused the neuron to activate. 

\begin{table}[]
\caption{Intra- and inter-cluster distance of embedded dataset examples averaged across all neurons in GPT2-small, comparing pre-MLP embeddings with neuron embeddings. Neuron embeddings on average result in denser clusters with better separation between clusters.}
\label{table:distances}
\vskip -0.1in
\begin{center}
\begin{small}
\begin{sc}
\begin{tabular}{@{}lcc@{}}
\toprule
Distance & Intra-cluster & Inter-cluster \\ \midrule
Pre-MLP  & 0.31          & 0.63          \\
Neuron   & 0.21          & 0.73          \\ \bottomrule
\end{tabular}
\end{sc}
\end{small}
\end{center}
\vskip -0.15in
\end{table}

\subsection{Training Sparse Auto-Encoders}

We provide a proof-of-concept showing how we can integrate information from neuron embeddings into the loss function when training Sparse Auto-Encoders (SAEs). We first train an MLP with one hidden layer containing $64$ neurons on the MNIST dataset \cite{mnist} for $3$ epochs, until the loss converges. We then experiment with training an SAE for the hidden layer with and without the neuron embedding loss term, and measure the effect on various evaluation metrics. 

Each SAE has a hidden dimension of $512$ (i.e., $8 \times$ the MLP hidden dimension), and we train them for one epoch over the training set. When incorporating the neuron embedding loss, we train for 200 steps ($\sim 40\%$ of an epoch) without the loss, then switch it on. This should allow the SAE neurons to stabilise, at which point the neuron embedding loss could be useful for pushing them to be more monosemantic. 

We provide a UI \footnote{\url{https://mechmnistic.streamlit.app}} which allows a user to examine any neuron in the MLP hidden layer or the SAE, and provides visualisations for interpreting the neuron's behaviour, which can be used to understand how the SAE neurons differ between the two models.

\begin{table}[]
\caption{
Evaluation metrics for SAEs trained with and without the neuron embedding (NE) loss. Accuracy loss is the absolute drop in accuracy after ablating the MLP neurons with the reconstructed activations from the SAE, from a starting accuracy of 94.0\%. 
}
\label{table:eval_stats}
\vskip -0.1in
\begin{center}
\begin{small}
\begin{sc}
\begin{tabular}{@{}lcccc@{}}
\toprule
          & MSE  & L1   & L0 / \% & Acc. Loss / \% \\ \midrule
Standard  & 0.33 & 2600 & 2.4     & 1.5            \\
+ NE Loss & 0.55 & 3300 & 9.1     & 4.6            \\ \bottomrule
\end{tabular}
\end{sc}
\end{small}
\end{center}
\vskip -0.15in
\end{table}

\begin{table}[]
\caption{Additional evaluation metrics for SAEs trained with and without the neuron embedding (NE) loss. 
Distances are measured on the neuron embeddings of each neuron's test set dataset examples, and size is the number of test set examples that induce a non-zero activation for a neuron. Dead refers to the percentage of neurons which don't activate for any example from the training dataset. 
}
\label{table:dist_stats}
\vskip -0.1in
\begin{center}
\begin{small}
\begin{sc}
\begin{tabular}{@{}lcccc@{}}
\toprule
          & Max Dist & Mean Dist & Size & Dead / \% \\ \midrule
Standard  & 0.45      & 0.21       & 7    & 23.8      \\
+ NE Loss & 0.28      & 0.08       & 32   & 3.7       \\ \bottomrule
\end{tabular}
\end{sc}
\end{small}
\end{center}
\vskip -0.15in
\end{table}

\begin{figure*}[t]
\centering
\includegraphics[width=0.75\textwidth]{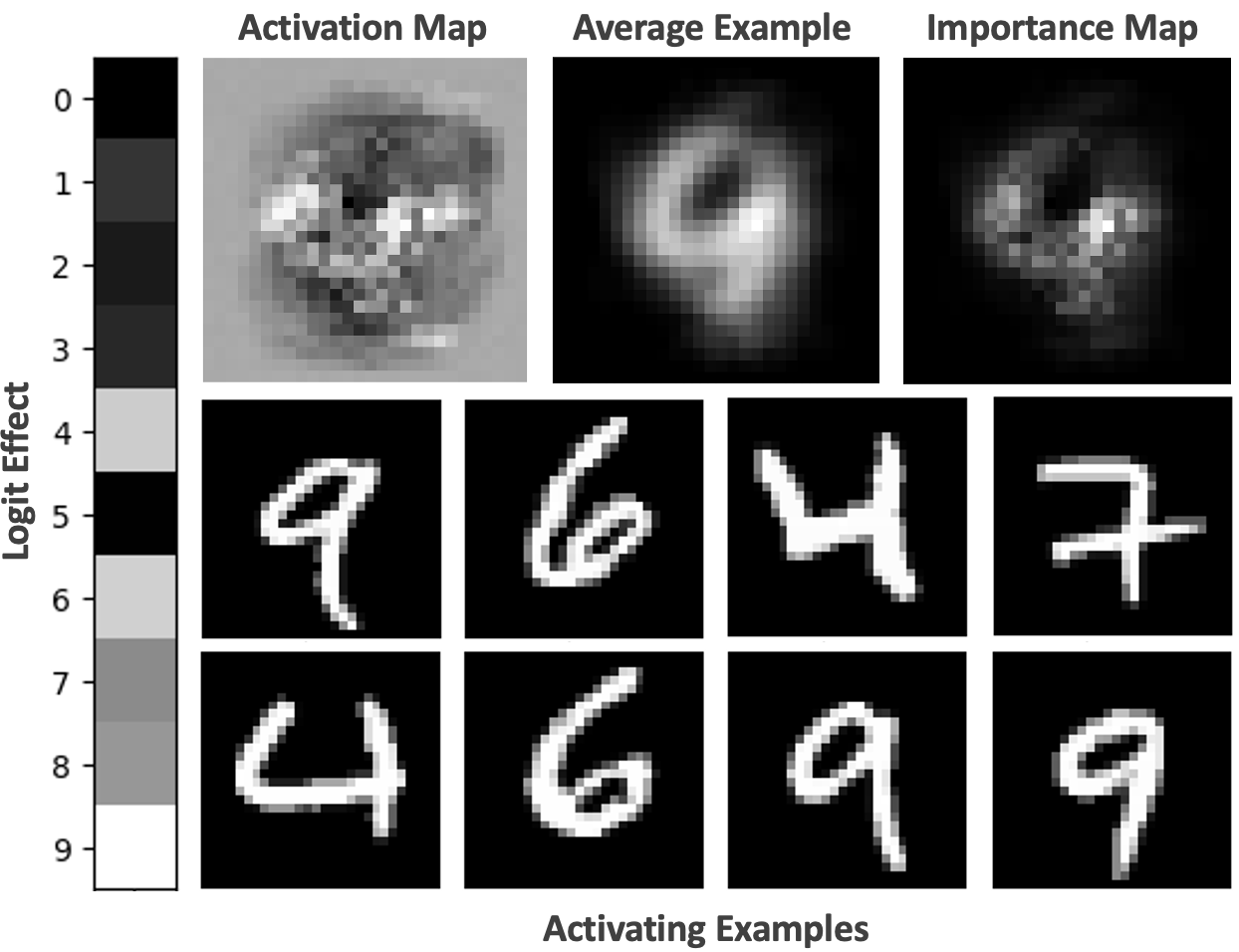}
\vskip -0.1in
\caption{Visualisations for an SAE neuron. The activation map shows the maximum magnitude of neuron activation for each pixel in the input, and the importance map is the average dataset example scaled by the activation map. 
}
\label{figure:sae}
\end{figure*}

Tables \ref{table:eval_stats} and \ref{table:dist_stats} show a suite of evaluation metrics for the SAEs trained with and without the neuron embedding (NE) loss. Table \ref{table:eval_stats} shows the typical evaluation metrics that measure reconstruction error and activation sparsity, measured on the held-out test data. Adding the NE loss increases the reconstruction error, reflected in the increased mean-squared error on the reconstructed activations and the significantly greater drop in absolute accuracy when ablating the MLP activations with the SAE reconstructions. It also decreases the sparsity, with the percentage of active neurons per input almost quadrupling from $2.4\%$ to $9.1\%$ and the L1 loss increasing as well. Note that decreasing sparsity is not intrinsically good or bad, but in the normal regime it is typically associated with decreased interpretability.

In contrast, Table \ref{table:dist_stats} shows improvements in the distance metrics. For each SAE neuron, we collected the test examples which caused a non-zero activation (up to a maximum of $100$ examples) and computed their neuron embeddings. We then measured the max and mean distance between points for each set of neuron embeddings, and took the median of these values over all neurons. We see significant decreases in the average max and mean distance between embeddings, which indicates potential improvements in monosemanticity. 

Interestingly, this is in spite of the decrease in sparsity and a corresponding increasing in the average number of activating examples (denoted ``size" in the table). We might expect that improvements in monosemanticity would come from neurons becoming more specific and responding to fewer examples, but in actuality we speculate that they have come from neurons moving to represent single, broader features, which respond to a wider variety of examples but look for the same information across all the examples. Anecdotally, we would tentatively suggest that this has corresponded to an improvement in SAE neuron interpretability, but we would recommend that readers investigate this for themselves by exploring the two different SAEs in the provided UI \footnote{\url{https://mechmnistic.streamlit.app/}}.

Figure \ref{figure:sae} shows an example neuron from the SAE trained with the NE loss that illustrates these broad, more general features that are learnt. The activation map shows the maximal activation that can be induced by each pixel, and the importance map is the element-wise product of the activation map and the average example. Note that this neuron had a single feature cluster with a mean and max distance of $0.01$ and $0.04$ between the neuron embeddings of the examples. 

The neuron appears to respond to lines or curves along the middle of the image, particularly towards the left and right edges. The logit effects show that the neuron increases the probability of predicting 9's, as well as 6's, 4's, and to a lesser extent 7's and 8's. This fits with the visual interpretation of the feature, and the randomly selected activating examples. This neuron visualisation demonstrates the style of feature which is commonly learned after including the NE loss, as well as how we can effectively understand a neuron's behaviour using some simple visualisations.

We also observe a more than $6 \times$ decrease in the percentage of dead neurons, which don't activate for any example in the training set, from $23.8\%$ to $3.7\%$. Dead neurons are a significant challenge when training SAEs, particularly as they are scaled up \cite{sonnet}. We haven't investigated why the NE loss causes such a significant decrease in the prevalence of dead neurons. Dead neurons shouldn't be directly penalised by the NE loss in theory, as it's only computed over the active neurons for a given input - in fact, increasing activation sparsity should decrease both the L1 and NE losses, but the NE loss seems to decrease sparsity instead. We speculate that encouraging neuron monosemanticity may force the SAE to utilise more of it's neurons and to learn more general features to avoid significant increases in the reconstruction error. Understanding the mechanism and effect of the NE loss in more detail would be a valuable direction for further research.


\section{Conclusion}

We presented neuron embeddings, and showed they can be used to effectively tackle neuron polysemanticity in a variety of ways. We used them to identify the distinct semantic behaviours of neurons in GPT2-small, and showed that they can capture the similarity between both individual examples and clusters of examples. By creating feature clusters to separate a neuron's dataset examples into their distinct behaviours we make it much easier to interpret the neuron, and also reduce the risk of running into the interpretability illusion by making it feasible to collect and summarise a wide variety of examples from across the activation spectrum. As dataset examples are an input into some automated interpretability techniques, applying feature clustering first could improve the results of these tools as well.

However, we note that our work doesn't consider features that may only emerge when considering several neurons together, which is a significant limitation. Future work could investigate using neuron embeddings and feature clusters in circuit analysis, or even look to extend the representation to multiple neurons. For example, it could be easier to understand how neurons co-activate to compensate for superposition by measuring which sub-neuron features activate together, rather than analysing neuron activation correlations directly as polysemanticity makes this very challenging. Additionally, in language models we only use the pre-MLP embedding of the token with the highest neuron activation. Combining the embeddings of multiple tokens, perhaps in proportion to their activation, may offer a better representation of the input.

We described how neuron embeddings can be used to measure neuron polysemanticity, which could be very useful for better evaluating SAEs. We also provided a proof-of-concept demonstrating how we can integrate information from neuron embeddings into the SAE loss. Applying this to a toy MLP model trained on MNIST showed several interesting effects, appearing to trade-off decreased reconstruction accuracy and activation sparsity for increased monosemanticity, as well as significantly decreasing the proportion of dead neurons. 

We note that this is early-stage research on a small toy model, so it remains unclear how these results would transfer to larger models. Applying neuron embeddings as an evaluation metric for SAEs trained on real-world language models, as well as experimenting with the neuron embedding loss when training such SAEs, would both be very interesting directions for future work.

\section*{Impact Statement}

This work presents a new method with applications in mechanistic interpretability of vision and language models. There are no specific ethical implications or societal consequences of this work that we feel need to be highlighted here.

\bibliography{main}
\bibliographystyle{icml2024}



\end{document}